\documentclass{article} 
\usepackage{iclr2022_conference,times}

\usepackage{amsmath,amsfonts,bm}

\def\eqref#1{equation~\ref{#1}}

\def\1{\bm{1}}

\DeclareMathAlphabet{\mathsfit}{\encodingdefault}{\sfdefault}{m}{sl}
\SetMathAlphabet{\mathsfit}{bold}{\encodingdefault}{\sfdefault}{bx}{n}

\usepackage{hyperref}
\usepackage{microtype}
\usepackage{graphicx}
\usepackage{amsmath}
\usepackage{amssymb}
\usepackage{amsfonts}
\usepackage{bm}
\usepackage{subfigure}
\usepackage{multirow}
\usepackage{dashrule}
\usepackage{subfigure}
\usepackage{booktabs} 
\usepackage{lipsum} 
\usepackage{stackengine}
\usepackage{colortbl}
\usepackage{adjustbox}
\usepackage{wrapfig}
\def\delequal{\mathrel{\ensurestackMath{\stackon[1pt]{=}{\scriptstyle\Delta}}}}
\definecolor{mygray}{gray}{.9}
\usepackage{url}
\definecolor{brown(traditional)}{rgb}{0.59, 0.29, 0.0}

\newtheorem{theorem}{Theorem}
\newtheorem{definition}{Definition}

\title{Towards Building A Group-based Unsupervised Representation Disentanglement Framework}

\author{Yang Tao$^{1}$\footnotemark[1]\thanks{Work done during internships at Microsoft Research Asia.}
, Xuanchi Ren$^{2}$
, Yuwang Wang$^3$\thanks{Corresponding author}  
, Wenjun Zeng$^4$
, Nanning Zheng$^1$ \\
$^1$Xi'an Jiaotong University,
$^2$HKUST, , $^3$Microsoft Research Asia, $^4$EIT \\
}

\iclrfinalcopy 
\begin{document}

\maketitle

\begin{abstract}
Disentangled representation learning is one of the major goals of deep learning, and is a key step for achieving explainable and generalizable models. A well-defined theoretical guarantee still lacks for the VAE-based unsupervised methods, which are a set of popular methods to achieve unsupervised disentanglement. The Group Theory based definition of representation disentanglement mathematically connects the data transformations to the representations using the formalism of \emph{group}. 
In this paper, built on the group-based definition and inspired by the \emph{n-th dihedral group}, we first propose a theoretical framework towards achieving \emph{unsupervised} representation disentanglement. We then propose a model, based on existing VAE-based methods, to tackle the unsupervised learning problem of the framework. In the theoretical framework, we prove three sufficient conditions on model, group structure, and data respectively in an effort to achieve, in an unsupervised way, disentangled representation per group-based definition. With the first two of the conditions satisfied and a necessary condition derived for the third one, we offer additional constraints, from the perspective of the group-based definition, for the existing VAE-based models.
Experimentally, we train 1800 models covering the most prominent VAE-based methods on five datasets to verify the effectiveness of our theoretical framework. Compared to the original VAE-based methods, these \emph{Groupified} VAEs consistently achieve better mean performance with smaller variances.
\end{abstract}
\section{Introduction}
\label{intro}
Learning independent and semantic representations of which individual dimension has interpretable meaning, usually referred to as disentangled representations learning, is critical for artificial intelligence research~\citep{bengio2013representation}. Such disentangled representations are useful for many tasks: domain adaptation~\citep{li2019cross,zou2020joint}, zero-shot learning~\citep{lake2017building}, and adversarial attacks~\citep{alemi2016deep}, etc. Intuitively, a disentangled representation should reflect the factors of variations behind the observed data of the world, and one latent unit is only sensitive to changes of an individual factor.

Due to the facts that obtaining the ground-truth labels requires significant human effort and humans can learn those factors unsupervisedly,
unsupervised representation disentanglement draws much attention from researchers recently.
A lot of methods are proposed base on some intuitions. Most of the state-of-the-art methods~\citep{higgins2016beta,burgess2018understanding, kim2018disentangling, chen2018isolating, kumar2017variational} are based on Variational Autoencoder (VAE)~\citep{kingma2013auto}. These methods are \emph{fully unsupervised} and can be applied to a variety of complex datasets~\citep{lee2020high}. However, these methods suffer from the unidentifiability problem~\citep{locatello2019challenging} due to a lack of theoretical guarantee. Another stream of works \citep{chen2016infogan, lin2020infogan, DS, lee2020high} leverage generative adversarial network (GAN) \citep{GAN} to achieve disentanglement but are not interpretable. In general, a well-defined theoretical guarantee is needed for those methods.

The research of symmetry in physics demonstrates that infinitesimal transformations that conform to some symmetry groups on physical objects can reflect their nature~\citep{anderson1972more, noether1915endlichkeitssatz}. Recently, inspired by this research on symmetry, \cite{higgins2018towards} proposed a group-based definition of disentangled representation. They argue that the symmetries, i.e., the transformations that change certain aspects of data and keep other aspects unchanged, ideally reflect the underlying data structure. The group-based definition is a formal and rigorous mathematical definition of faithful and, ideally, interpretable representation of the generative
factors of data, which is widely accepted~\citep{greff2019multi, mathieu2019disentangling, khemakhem2020variational}. Subsequently, due to the fact that the definition is defined by the world state (i.e., Ground Truth) and based on the assumption ~\citep{caselles2019symmetry} that this definition should be useful for downstream tasks such as a Reinforcement Learning, ~\cite{caselles2019symmetry}, ~\cite{quessard2020learning}, ~\cite{painter2020linear} propose environment-based (to provide world state) methods to learn such disentangled representations in Reinforcement Learning settings. These inspire us to ask the following question: how would the definition benefit \emph{unsupervised} representation disentanglement, and how to learn such a disentangled representation conforming to the definition in the setting of \emph{unsupervised} representation learning?

In Group Theory\footnote{We assume some basic familiarity with the fundamentals of Group Theory and Group Representation Theory. Please refer to Appendix A for some basic concepts.}, the \emph{n-th dihedral group}~\citep{judson2020abstract} is a set of all permutations of polygons vertices, forming a permutation group under the operation of composition \citep{miller1973symmetry}. The generators in an \emph{n-th dihedral group}, i.e., flip and rotation, can be regarded as the disentangled factors and also transformations. In this paper, inspired by the \emph{n-th dihedral group}, we answer the above questions and address the challenge by proposing a theoretical framework to make the definition practically applicable for \emph{unsupervised} representation disentanglement. We then propose a model to tackle the learning problem of the framework and verify its effectiveness.  We theoretically prove in Section \ref{non_id_bs} the three sufficient conditions towards achieving disentangled representation per group-based definition, which are referred to as model, group structure, and data constraint, respectively. With these conditions, we offer additional constraints from the perspective of the definition. The additional constraints encourage existing VAE-based models to satisfy the symmetry requirement that comes from the nature of factors. Finally, we provide a learning model based on the existing VAE-based methods in an effort to fulfill the three conditions (with the model and group structure constraint and a \textit{necessary} condition for the data constraint satisfied). As an intuitive understanding, we introduce the additional constraints to reorganize the latent space to restrict its symmetry in an unsupervised way. These additional constraints indeed narrow down the solution space of VAE-based models. Detailed discussion in Sec. \ref{Sec:latent_space}.
Our model consistently achieves statistically better performance in prominent metrics (higher means and lower variances) than corresponding existing VAE-based models on five datasets, demonstrating that the group-based definition together with our proposed framework further encourages disentanglement.

Our main contributions are summarized as follows:
\begin{itemize}
\item To our best knowledge, we are the first to provide a theoretical framework to make the formal group-based mathematical definition of disentanglement practically applicable to \emph{unsupervised} representation disentanglement.
\item Our theoretical framework provides additional constraints from the perspective of group-based definition for the existing VAE-based methods.
\item We propose a learning model of the framework by deriving and integrating additional loss into existing VAE-based models, in an effort to make the learned representation conform to the group-based definition without relying on the environment (as done in \cite{caselles2019symmetry, quessard2020learning, painter2020linear}).
\end{itemize}

\section{Related Works}
Different definitions have been proposed for disentangled representation~\citep{bengio2013representation, higgins2018towards, suter2019robustly}. However, only the group-based definition proposed by~\cite{higgins2018towards} focuses on the disentangled representation itself and is mathematically rigorous, which is well accepted~\citep{caselles2019symmetry,quessard2020learning, painter2020linear, TopologDefects}. Nevertheless, ~\cite{higgins2018towards} 
do not propose a specific learning method based on their definition. Before this rigorous definition was proposed, there had been some success in identifying generative factors in static datasets (without interaction with environment), e.g., $\beta$-VAE~\citep{higgins2016beta}, Anneal-VAE~\citep{burgess2018understanding}, $\beta$-TCVAE~\citep{chen2018isolating}, and FactorVAE~\citep{kim2018disentangling}.   More recent works \citep{srivastava2020improving, shao2020controlvae, kim2019bayes, lezama2018overcoming, rezende2018taming} also do not consider the group-based definition. Therefore, how group-based definition will facilitate these methods is still an open question.
Besides, all these works suffer from the unidentifiability problem~\citep{locatello2019challenging}, which is a challenging problem in this literature. From group-based definition, our framework points out that, the unidentifiability problem could be solved once the data constraint is satisfied. However, in this work, we can only get a necessary condition for data constraint, and we still can not solve this challenging problem.


As pointed out in ~\cite{quessard2020learning}, it is not straightforward to reconcile the probabilistic inference methods with the group-based definition framework.  ~\cite{caselles2019symmetry}, ~\cite{quessard2020learning}, ~\cite{painter2020linear} leverage the interaction with the environment (assuming it is available) as supervision instead of minimizing the total correlation as the VAE-based methods do. Consequently, the effectiveness of these methods is limited to the datasets with the environment available. \emph{Our framework learns a representation conforming to the group-based definition without relying on the environment.} ~\cite{pfau2020disentangling} propose a non-parametric method to unsupervisedly learn linear disentangled planes in data manifold under a metric. However, as pointed out by the authors, the method does not generalize to held-out data and performs poorly when trying to disentangle directly from pixels.

To summarize, the existing probabilistic inference methods lack theoretical support, while the application scope of existing methods based on the group-based mathematical definition~\cite{higgins2018towards} is very limited.  To the best of our knowledge, \emph{our work is the first to reconcile the probabilistic generative methods with the inherently deterministic group-based definition framework of ~\cite{higgins2018towards}.}

\section{The Group-based Framework For Unsupervised Representation Disentanglement}
\label{method}
Our goal is to explore the benefit of the group-based definition for \emph{unsupervised} representation disentanglement and learn such a disentangled representation. The background of the group-based definition is provided in Section~\ref{dis_def}. Section~\ref{non_id_bs} presents the theoretical framework towards achieving \emph{unsupervised} disentanglement, in which we derive three sufficient conditions on the model, group structure, and data, respectively. The conditions on the model and group structure provide additional constraints for the existing VAE-based models.

\subsection{Group-based Definition}
\label{dis_def}
We briefly review the group-based definition of disentangled representation~\cite{higgins2018towards}.
Considering a group $G$ acting on world state space $W$ (can be understood as ground-truth) of data space $O$ and representation space $Z$ via \emph{group action} $\cdot _W$ and \emph{group action} $\cdot _Z$  respectively. For a mapping $f = b\circ h$, where $b$ and $h$ denote the data generative process and encoding, we state: the mapping $f$ is \emph{equivariant} between the actions on $W$ and $Z$ if
\begin{equation}
    g\cdot f(w) = f(g\cdot w),\; \forall g \in G, \; \forall w \in W.
\label{equ:equivariant}
\end{equation}
\begin{definition}
\vspace{-1.5em}
\label{def1}
Assume $G$ can be decomposed as $G= G_1\times G_2\times \dots \times G_m$. The set $Z$ is disentangled with respect to $G$ if: $(i)$ \emph{group action} of $G$ on $Z$ exits. $(ii)$ the mapping $f$ is equivariant between the actions on $W$ and $Z$. $(iii)$ There is a decomposition $Z = Z_1\times Z_2 \times\dots \times Z_m$ such that each $Z_i$ is affected only by the corresponding $G_i$.
\end{definition}

It is challenging to apply the group-based definition to an \emph{unsupervised} disentanglement setting in practice because the definition refers to the world state space $W$, the group action of $G$ on $W$, and mapping $b$ which are typically inaccessible in practice. We tackle the challenge by re-framing the definition in a new framework in the following section.

\subsection{Proposed Theoretical Framework}
\label{non_id_bs}


Since when the representation is disentangled, one latent unit in the representation space is only sensitive to changes of an individual generative factor, we make the following assumptions: $G$ is a direct product of $m$ \emph{cyclic groups} (as suggested by~\cite{higgins2018towards} and for simplicity): $G = (\mathbb{Z}/n\mathbb{Z})^m = \mathbb{Z}/n\mathbb{Z}\times \mathbb{Z}/n\mathbb{Z}\times \dots \times \mathbb{Z}/n\mathbb{Z}$, where $n$ is the assumed total number of possible values for a factor and $m$ is the total number of factors; we further assume $Z$ is a set with the same elements in $G$. Therefore, the \emph{group actions} of $G$ on $Z$ can be set to be element-wise addition, i.e., $g\cdot z = \overline{g + z},\forall z\in Z, g \in G$. For the \emph{generator} of dimension $i$ of $G$, $g_i = (0,\dots, \overline{1},\dots,0)$, $g_i$ only affects the $i$-th dimension of $z$ by $\overline{g_i + z}$. In addition, the action of each generator $g_i$ on $w$ only affects a single dimension of $w$.

As we can seen from Equation \ref{equ:equivariant} above, the group action is defined on $w$, which is often not accessible, making it difficult to apply the definition in practice. Therefore, for the \emph{unsupervised} setting, we would like to use permutations on the data space $O$ (which only provides data without labels) to substitute the group actions on $W$. Specifically, inspired by the \emph{n-th dihedral group}~\citep{dummit1991abstract}, we construct a \emph{permutation group} $\Phi$, serving the role of an ``agent'' of $G$. The actions of $G$ on $W$ can be performed by $\varphi_g\in \Phi$ on $O$, which can be formulated as
\begin{equation}
f(g\cdot w) = h(\varphi_g\cdot b(w)) = h(\varphi_g\cdot o), \forall w \in W, g \in G,
\label{equ:agent}
\end{equation}
where $o$ denotes the data (e.g., image) corresponding to the world state $w$ through the mapping function $b$. If the above equation holds, we state that the ``agent'' permutation group $\Phi$ exists. We first give the conditions for the existence of this ``agent'' permutation group $\Phi$, then derive the additional condition to achieve such disentanglement. We accomplish these two objectives in Theorem \ref{t1} with the proof provided in Appendix B. Theorem \ref{t1} states that a \emph{general permutation group} $\Phi$ on $O$ can serve as an agent group (\emph{agent group exists}) if and only if both $(\mathtt{i})$ and $(\mathtt{ii})$ are satisfied. If the agent group exists, and its \emph{permutations} (actions on $O$) can be defined by an autoencoder-like model as shown in the equation in $(\mathtt{iii})$, then $Z$ is disentangled with respect to $G$.

\begin{theorem}
\vspace{-0.6em}
\label{t1}
For the group $G=(\mathbb{Z}/n\mathbb{Z})^m$, a permutation group $\Phi$ on $O$, a representation space $Z$, a World State space $W$, and mapping $b$ and $h$, Equation \ref{equ:agent} holds if and only if $(\mathtt{i})$ $\Phi$ is isomorphic to $G$, and $(\mathtt{ii})$ For each \emph{generator} of dimension $i$ of $G$, $g_i$, there exists a $\varphi_i \in \Phi, i=1,\dots,m$, such that $\varphi_i\cdot b(w) = b(g_i\cdot w),\; \forall w\in W$, and $\varphi_i$ is a generator of $\Phi$; Further, if Equation \ref{equ:agent} holds and $(\mathtt{iii})$ $\varphi_g\cdot b(w) = h^{-1}(g\cdot f(w))\;\forall w\in W, \varphi_g \in \Phi$, then $Z$ is disentangled with respect to $G$, where $\varphi_g$ is the corresponding element in $\Phi$ of $g$ under the isomorphism.
\vspace{-0.6em}
\end{theorem}

In Theorem \ref{t1}, $(\mathtt{i})$ states that the relation between the elements (i.e., group structure) is preserved between $\Phi$ and $G$, and we denote it as the \emph{group structure constraint}; $(\mathtt{ii})$ actually indicates a data constraint that all variations in the data can be generated by compositions of some basic \emph{permutation generators} $\{\varphi_i\}_{i = 1,\dots,m}$. We denote it as the \emph{data constraint}; $(\mathtt{iii})$ states that the \emph{permutations} in the agent group $\Phi$ are defined by encoding, action, and decoding, which is referred to as the \emph{model constraint}. Note that in Theorem \ref{t1}, only the data constraint refers to the world state $w$.

Here is a sketch of the proof: \emph{data constraint} is a special case of Equation \ref{equ:agent} for a \emph{generator}, and \emph{group structure constraint} is a relation-preserving constraint on compositions of \emph{generators}, and satisfying both constraints will thus result in that Equation \ref{equ:agent} holds for any general element in $\Phi$, and vice versa. Moreover, we can derive Equation \ref{equ:equivariant} for disentanglement when combining the \emph{model constraint} and Equation \ref{equ:agent}. 

The model constraint specifies the way to permute the data. When the data is permuted, its world state changes. Therefore, how the world states transit between each other is modeled by the \emph{model constraint} applied on the data. The isomorphism between $\Phi$ and $G$ ensures that the world state space $W$ and data space $O$ have the same symmetry. In this way, the model applied on the data learns the transition of the world states. Note that we aim to bring this group-based definition, which requires ground truth by default, into the \emph{unsupervised} setting. Now only the data constraint refers to the world states, and it seems almost impossible to derive a sufficient condition for it without the labels. We thus make a trade-off in which we use a necessary condition in the next section.

\section{Groupified VAE: a learning method of the framework}
\label{relation}
Let's look closer into the three constraints, respectively. Firstly, we consider the model constraint, $\varphi_g\cdot o = h^{-1}(g\cdot h(o))\;\forall o\in O, \varphi_g \in \Phi$, which suggests that the action of $\Phi$ on $O$ can be implemented using an autoencoder-like network that performs encoding, action on its representation space, and decoding.  Given an autoencoder-like network with an encoder $h$ and a decoder $d$, since $d$ is approximately the inversion of $h$, the \emph{model constraint} can be formulated as
\begin{equation}
    \varphi_g\cdot o = h^{-1}(g\cdot h(o)) \delequal d(g\cdot h(o)), \forall o \in O, g \in G,
    \label{equ:groupaction}
\end{equation}

together with further implementation of $\Phi$ as described in Section \ref{groupifiable}, the \emph{model constraint} can be fulfilled. Secondly, The \emph{data constraint} requires that all variations in the data can be generated by compositions of some basic \emph{permutations generators}. Previous VAE-based works~\citep{higgins2016beta,burgess2018understanding, kim2018disentangling, chen2018isolating, kumar2017variational} aim to generate the data with independent generative factors, which is in line with the \emph{data constraint}.
Intuitively, if the VAE-based model can generate the data from statistical independent basic latent units and each unit corresponds to the basic \emph{permutation generator}, the \emph{data constraint} may be satisfied. Based on the intuition above, we prove that if the world state is independently sampled per dimension, the minimization of total correlation is a necessary condition for the \emph{data constraint}  (see Appendix E). Therefore, we can leverage existing VAE-based models to fulfill the \emph{data constraint} to some extent for the \emph{unsupervised} setting. Lastly, to satisfy the \emph{group structure constraint}, we derive a self-supervised \emph{Isomorphism Loss} which can be incorporated into the VAE-based model as described in Section \ref{isom_loss}. 

\begin{figure*}[t]
\centering
\includegraphics[width=\linewidth]{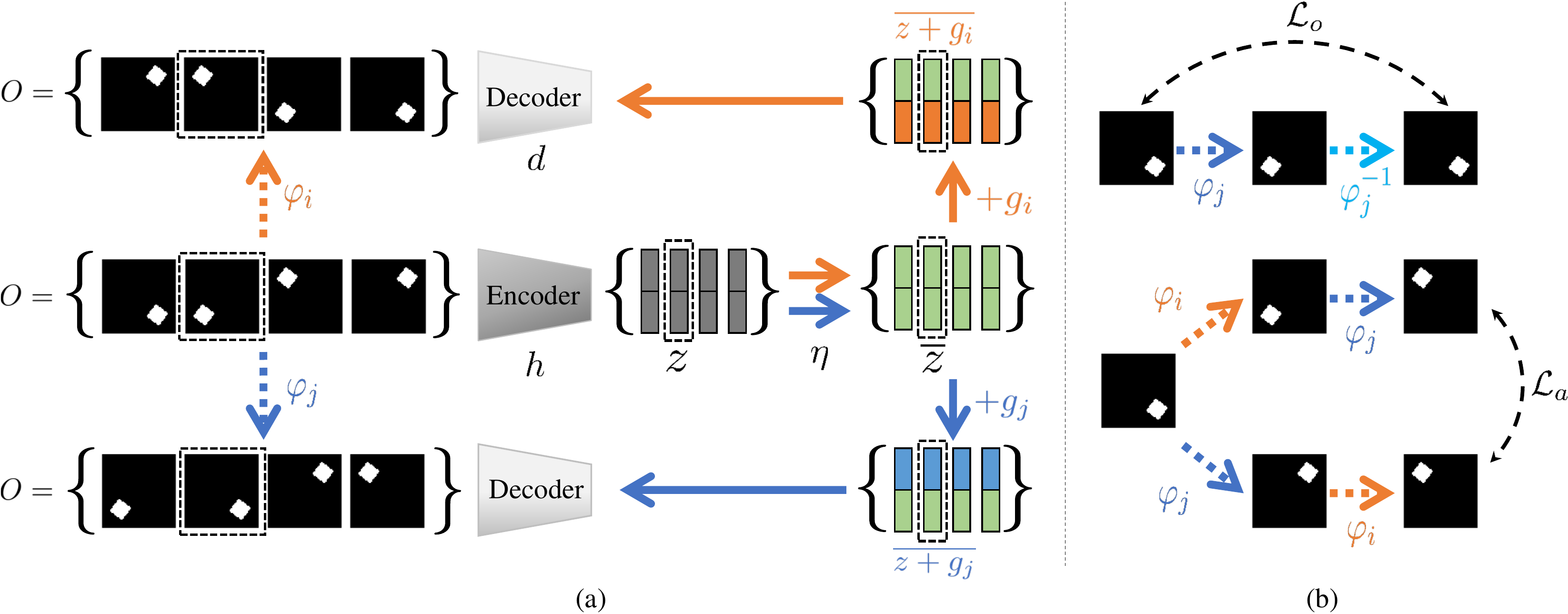}
\vspace{-1.7em}
\caption{Overview of the implementation (\emph{Groupified} VAE). (a) Illustration of \emph{permutation group} $\Phi = \{\varphi_g|g\in G\}$ defined on a VAE-based model, where $G = (\mathbb{Z}/n\mathbb{Z})^m$. The \emph{generators} $\varphi_i, \varphi_j\in \Phi$ are permutations on $O$. Specifically, when optimized, $\varphi_i$ and $\varphi_j$ are horizontal and vertical movements. $\varphi_i$ is defined as the solid orange arrows illustrate: encode an image $o$ to representation $z$, perform $\eta$ on $z$ to get $\overline{z}$, add $g_i$ to $\overline{z}$, and decode back to the image. This process can be regarded as an exchange of images in dataset (permutation), as the dashed orange arrow shows. These permutations form a group $\Phi$. (b) The Isomorphism Loss, which guarantees that $\Phi$ is isomorphic to $G$, includes Abel Loss $\mathcal{L}_a$ constraining the commutativity, and Order Loss $\mathcal{L}_o$ constraining the cyclicity.}
\vspace{-1.5em}
\label{fig:frame_work}
\end{figure*}

\subsection{Implementation of Group \texorpdfstring{$\Phi$}{Phi}}
\label{groupifiable}

The key is to implement the \emph{group actions} of $G$ on $Z$ into the VAE-based models, we need to map the representation $z$ to a group that is isomorphic to $G$ (cyclic representation space).
Therefore, we construct a function $\eta$ to achieve this mapping.  Moreover, this mapping is required to be differentiable, in order for back-propagation to be adopted for optimization. According to Group Theory, there is an isomorphism between $G$ and the \emph{n-th root unity group}: $\{\exp((2\pi i z)/n)|z\in \mathbb{Z}^m\}$, where $n,m$ are the same as in $G$. Therefore, the representation $z$  can be mapped to $\overline{z}$ by the function $\eta$ as $\overline{z}=\eta(z)=\exp((2{\pi}iz)/n)$ (see Figure \ref{fig:frame_work} (a)). However, $\overline{z}$ can not be mapped to directly as it has complex numbers, but we can use Euler's formula: $\exp((2\pi i z)/{n}) = \sin ((2\pi z)/{n}) + i\cos ((2\pi z)/{n})$ to map $z$ to its real and imaginary part, i.e.,  vector $\sin ((2\pi z)/{n})$  and $\cos ((2\pi z)/{n})$. The two vectors are concatenated and fed to the decoder. 

For ease of implementation, the \emph{permutation group} $\Phi$ can be approximately generated by compositions of \emph{generators}, i.e., $\Phi = <\varphi_1,\varphi_2,\dots,\varphi_m>$. Recall that the \emph{generator} $\varphi_i$ of group $\Phi$ is defined as $\varphi_i\cdot o = d(g_i\cdot h(o)) = d(\overline{h(o)+g_i}), \forall o \in O$, where $g_i$ is \emph{generator} of dimension $i$ in $G$, as shown in Figure~\ref{fig:frame_work} (a). For $\varphi_i$, we implement $g_i\cdot h(o)$ by adding $1$ (without loss of generality) to the $i$-th dimension of $h(o)$, then make it cyclic by function $\eta$. Similarly, for $\varphi_i^{-1}$, we add the value of $n-1$.

\subsection{Implementation of the Isomorphism}
\label{isom_loss}
In this section, to satisfy the \emph{group structure constraint} (isomorphism), we derive two equivalent constraints, which are then converted into an \emph{Isomorphism Loss $\mathcal{L}_I$}.
Many groups are uniquely determined by the properties of the \emph{generators}, e.g., group $G = <a,b|a^2=b^2=e, ab=ba>$. In addition, since the group $\Phi$ is \emph{isomorphic} to G, $\Phi$ is also expected to be commutative and cyclic.
In light of this, we derive two constraints on \emph{generators} that  are equivalent to the isomorphism condition, as described in Theorem \ref{t2}. Please refer to Appendix C for the proof.

\begin{theorem}
\vspace{-0.5em}
\label{t2}
The defined permutation group $\Phi = <\varphi_1, \varphi_2,\dots, \varphi_m>$ is isomorphic to  $G=(\mathbb{Z}/n\mathbb{Z})^m$ if and only if: $(i)$ for $\forall$ generators $\varphi_i, \varphi_j \in \Phi, 1 \leq i,j \leq m$, we have $\varphi_i\varphi_j = \varphi_j \varphi_i$, and $(ii)$ $\forall \varphi_i \in \Phi, 1 \leq i \leq m$, we have $\varphi_i^n = e$, where $e$ is the identity element of group $\Phi$.
\vspace{-0.5em}
\end{theorem}

The first constraint requires the group $\Phi$ to be an \emph{abelian group}~\citep{judson2020abstract}. Therefore, we denote it as Abel constraint and the loss derived from it as the Abel Loss $\mathcal{L}_a$. The second is a constraint on the \emph{order} of elements. We thus denote it as the Order constraint and the loss derived from it as the Order Loss $\mathcal{L}_o$. See Appendix F for a more detailed implementation.

\textbf{Abel Loss.}
For the Abel constraint: $\forall \varphi_i, \varphi_j \in \Phi, 1 \leq i,j \leq m$, we have $\varphi_i\varphi_j = \varphi_j\varphi_i$. We minimize $ \|\varphi_i\cdot(\varphi_j\cdot o) - \varphi_j\cdot(\varphi_i\cdot o)\|, \forall o \in O$ to meet the Abel constraint, as shown in Figure \ref{fig:frame_work} (b). The Abel Loss is the sum of the losses of all combinations of two \emph{generators}. Denote the set of combinations as $C = \{(i,j)|1 \leq i,j \leq m\}$. The Abel Loss is defined as follows
\begin{equation}
    \mathcal{L}_{a}  = \sum_{o\in O} \sum_{(i,j)\in C} \| \varphi_{i}\cdot (\varphi_{j}\cdot o) - \varphi_{j}\cdot (\varphi_{i}\cdot o)\|.
    \label{equ:ab_ba_main}
\end{equation}

\textbf{Order Loss.}
For the Order constraint: $\forall \varphi_i \in \Phi, 1 \leq i \leq m$, we have $\varphi_i^n = e$, where $e$ is the \emph{identity element} in group $\Phi$ (identity mapping). Note that with $n$ times composition of $\varphi_i$, it is difficult for the gradient to back-propagate. We thus use an approximation that uses $2$ times of composition instead. When the autoencoder can do the reconstruction well, this approximation holds, see appendix E for details.
Similar to Abel Loss, we minimize $\|\varphi_i\cdot (\varphi_i^{n-1}\cdot o) - o\|, \forall o \in O$ to satisfy the Order constraint. The whole process is illustrated in Figure \ref{fig:frame_work} (b). 
However, the equation is not symmetrical and leads to bias. Therefore, we use the following symmetrical form instead:
\begin{equation}
    \mathcal{L}_{o}  = \sum_{o\in O} \sum_{1\leq i\leq m} \left(\| \varphi_i\cdot(\varphi_i^{n-1}\cdot o) - o\|
    + \|\varphi_i^{n-1}\cdot(\varphi_i\cdot o) -o\|\right).
    \label{equ:an_e_main}
\end{equation}

With the above two loss functions optimized, the isomorphism condition is satisfied, which can be illustrated by Theorem~\ref{t3}. Please refer to Appendix D for the proof.

\begin{theorem}
\vspace{-0.5em}
\label{t3}
The following two conditions are equivalent: $(i)$ $\forall \varphi_i, \varphi_j \in \Phi, 1 \leq i,j \leq m$, we have $\varphi_i\varphi_j = \varphi_j \varphi_i$ and $\forall \varphi_i \in \Phi, 1 \leq i \leq m$, we have $\varphi_i^n = e$ $(ii)$ the Abel Loss function (Equation~\ref{equ:ab_ba_main}) and the Order Loss  function (Equation~\ref{equ:an_e_main}) are optimized.
\vspace{-0.5em}
\end{theorem}

Since the Abel Loss and Order Loss are equally important for satisfying the isomorphism condition, we assign equal weight to them. Thus, the \textbf{Isomorphism Loss} is $\mathcal{L}_{I} = \mathcal{L}_{o} + \mathcal{L}_{a}$. With the implementation of group $\Phi$, the \emph{model constraint} is satisfied. We optimize the Isomorphism Loss to satisfy the \emph{group structure constraint}. To further satisfy the data constraint to some extent as described in Section \ref{relation}, we leverage VAE-based models and optimize their original loss (that minimizes the total correlation), denoted as $\mathcal{L}_{VAE}$. Therefore, the \textbf{Total Loss} is $\mathcal{L}  = \mathcal{L}_{VAE} + \gamma_I\mathcal{L}_{I}$, where $\gamma_I$ is the weight of Isomorphism Loss. We denote the above VAE-based implementation as \emph{Groupified} VAE.

\section{Experiments}
We first verify the effectiveness of \emph{Groupified} VAE quantitatively in learning disentangled representations on several datasets and several VAE-based models. Then, we show its effectiveness qualitatively on two typical datasets. After that, we perform a case study on the dSprites dataset to analyze the effectiveness, and conduct ablation studies on the losses and hyperparameters. For the performance comparison of two downstream tasks (abstract reasoning ~\cite{van2019disentangled} and fairness evaluation ~\cite{locatello2019fairness}), and more comprehensive results, please see Appendix I.

\subsection{Datasets and Baseline Methods}
To evaluate our method, we consider several datasets: dSprites~\citep{higgins2016beta}, Shapes3D~\citep{kim2018disentangling}, Cars3D~\citep{car3d}, and the variants of dSprites introduced by Locatello et al.~\citep{locatello2019challenging}: Color-dSprites and Noisy-dSprites. Please refer to Appendix G for the details of the datasets.

We choose the following four baseline methods as representatives of the existing VAE-based models, which are denoted as Original VAEs. We verify the effectiveness of our implementation based on those methods. \textbf{$\bm{\beta}$-VAE}~\citep{higgins2016beta} introduces a hyperparameter $\beta$ in front of the KL regularizer of the VAE loss. It constrains the VAE information capacity to learn the most efficient representation. \textbf{AnnealVAE}~\citep{burgess2018understanding} progressively increases the bottleneck capacity so that the encoder learns new factors of variation while retaining disentanglement in previously learned factors. \textbf{FactorVAE}~\citep{burgess2018understanding} and \textbf{$\beta$-TCVAE}~\citep{chen2018isolating} both penalize the total correlation~\citep{watanabe1960information}, but estimate it with adversarial training~\citep{nguyen2010estimating,sugiyama2012density} and Monte-Carlo estimator respectively.

\begin{table*}[htbp!]
\centering

\resizebox{\textwidth}{!}{
\begin{tabular}{c c>{\columncolor{mygray}}c c>{\columncolor{mygray}}c c>{\columncolor{mygray}}c c>{\columncolor{mygray}}c}
\toprule
 \multirow{2}*{dSprits}&  \multicolumn{2}{c}{DCI} & \multicolumn{2}{c}{BetaVAE} & \multicolumn{2}{c}{MIG} & \multicolumn{2}{c}{FactorVAE}\\ 
\cmidrule(lr){2-9}
 & Original & Groupified & Original & Groupified & Original & Groupified & Original & Groupified \\
 \midrule
 $\beta$-VAE &  $ 0.23 \pm 0.10 $  & $ \bm{0.46} \pm \bm{0.085} $ & $ 0.75 \pm 0.083 $  & $ \bm{0.86} \pm \bm{0.051} $  & $ 0.14 \pm 0.097 $  & $ \bm{0.37} \pm \bm{0.089} $  & $ 0.51 \pm 0.098 $  & $ \bm{0.63} \pm \bm{0.089} $  \\
 AnnealVAE  & $ 0.28 \pm 0.10 $  & $ \bm{0.39} \pm \bm{0.056} $  & $ 0.84 \pm 0.050 $  & $ \bm{0.87} \pm \bm{0.0067} $ & $ 0.23 \pm 0.10 $  & $ \bm{0.34} \pm \bm{0.061} $  & $ \bm{0.70} \pm 0.094 $  & $ 0.68 \pm \bm{0.058} $  \\
 FactorVAE  & $ 0.38 \pm 0.10 $  & $ \bm{0.41} \pm \bm{0.074} $  & $ \bm{0.89} \pm 0.040 $  & $ \bm{0.89} \pm \bm{0.020} $ & $ 0.27 \pm 0.092 $  & $ \bm{0.31} \pm \bm{0.061} $  & $ 0.74 \pm 0.068 $  & $ \bm{0.75} \pm 0.075 $   \\
 $\beta$-TCVAE  & $ 0.35 \pm \bm{0.065} $  & $ \bm{0.36} \pm 0.11 $  & $ 0.86 \pm \bm{0.026} $  & $ \bm{0.861} \pm 0.038 $  & $  0.17 \pm 0.067 $  & $ \bm{0.24} \pm 0.093 $  & $ 0.68 \pm 0.098 $  & $ \bm{0.70} \pm 0.098 $ \\
 \bottomrule
\end{tabular}}

\vspace{1em}
\resizebox{\textwidth}{!}{
\begin{tabular}{c c>{\columncolor{mygray}}c c>{\columncolor{mygray}}c c>{\columncolor{mygray}}c c>{\columncolor{mygray}}c}
\toprule
  \multirow{2}*{Cars3d}&  \multicolumn{2}{c}{DCI} & \multicolumn{2}{c}{BetaVAE} & \multicolumn{2}{c}{MIG} & \multicolumn{2}{c}{FactorVAE}\\ 
\cmidrule(lr){2-9}
 & Original & Groupified & Original & Groupified & Original & Groupified & Original & Groupified \\
 \midrule
 $\beta$-VAE & $ 0.18 \pm 0.059 $  & $ \bm{0.24} \pm \bm{0.041} $  & $ 0.99 \pm 1.6e-3 $  & $ \bm{1.0} \pm \bm{0.0} $ & $ 0.071 \pm 0.032 $  & $ \bm{0.11} \pm 0.032 $  & $ 0.81 \pm 0.066 $  & $ \bm{0.93} \pm \bm{0.034} $  \\
 AnnealVAE  & $ 0.22 \pm 0.046 $  & $ \bm{0.25} \pm \bm{0.046} $  & $ \bm{0.99} \pm 4e-4 $  & $ \bm{0.99} \pm \bm{1.5e-4} $ & $ 0.074 \pm 0.016 $  & $ \bm{0.10} \pm \bm{0.014} $  & $ 0.82 \pm 0.062 $  & $\bm{0.87} \pm \bm{0.028} $  \\
 FactorVAE  & $ 0.21 \pm 0.054 $  & $ \bm{0.25} \pm \bm{0.040} $  & $ 0.99 \pm 1e-4 $  & $ \bm{1.0} \pm \bm{0.0} $ & $ 0.098 \pm 0.027 $  & $ \bm{0.11} \pm 0.033 $  & $ 0.90 \pm 0.039 $  & $ \bm{0.93}\pm \bm{0.034} $   \\
 $\beta$-TCVAE  & $ 0.24 \pm 0.049 $  & $ \bm{0.26} \pm \bm{0.046} $  & $ \bm{1.0} \pm \bm{0.0} $  & $ \bm{1.0} \pm \bm{0.0} $ & $ 0.10 \pm 0.021 $  & $ \bm{0.11} \pm 0.033 $  & $ 0.88 \pm 0.040 $  & $\bm{0.93} \pm \bm{0.034} $  \\
 \bottomrule
\end{tabular}}

\vspace{1em}
\resizebox{\textwidth}{!}{
\begin{tabular}{c c>{\columncolor{mygray}}c c>{\columncolor{mygray}}c c>{\columncolor{mygray}}c c>{\columncolor{mygray}}c}
\toprule
  \multirow{2}*{Shapes3d}&  \multicolumn{2}{c}{DCI} & \multicolumn{2}{c}{BetaVAE} & \multicolumn{2}{c}{MIG} & \multicolumn{2}{c}{FactorVAE}\\ 
\cmidrule(lr){2-9}
 & Original & Groupified & Original & Groupified & Original & Groupified & Original & Groupified \\
 \midrule
 $\beta$-VAE & $ 0.44 \pm 0.176 $  & $ \bm{0.56} \pm\bm{0.10} $  & $ 0.91 \pm 0.072 $  & $ 0.90 \pm \bm{0.045} $ & $ 0.28 \pm 0.18 $  & $ \bm{0.42} \pm \bm{0.15} $  & $ \bm{0.82} \pm 0.098 $  & $ \bm{0.82} \pm \bm{0.043} $  \\
 AnnealVAE  & $ 0.52 \pm \bm{0.051} $  & $ \bm{0.60} \pm 0.078 $  & $ 0.82 \pm \bm{0.076} $  & $ \bm{0.89} \pm 0.086 $ & $ 0.48 \pm 0.047 $  & $ \bm{0.50} \pm 0.052 $  & $ 0.75 \pm 0.074 $  & $ \bm{0.83} \pm \bm{0.066} $  \\
 FactorVAE  & $ 0.47 \pm 0.10 $  & $ \bm{0.49} \pm \bm{0.065} $  & $ \bm{0.86} \pm \bm{0.055} $  & $ 0.80 \pm 0.075 $ & $ 0.33 \pm 0.13 $  & $ \bm{0.43} \pm \bm{0.11} $  & $ \bm{0.81} \pm \bm{0.056} $  & $ 0.79 \pm 0.066 $   \\
 $\beta$-TCVAE  & $ 0.66 \pm 0.10 $  & $ \bm{0.72} \pm \bm{0.061} $  & $ \bm{0.97} \pm \bm{0.039} $  & $ 0.96 \pm 0.042 $  & $ 0.40 \pm 0.18 $  & $ \bm{0.47} \pm \bm{0.090} $  & $ 0.89 \pm 0.064 $  & $ \bm{0.90} \pm \bm{0.046} $ \\
 \bottomrule
\end{tabular}}

\caption{Performance (mean $\pm$ std) on different datasets and different models with different metrics. We evaluate $\beta$-VAE, AnnealVAE, FactorVAE, and $\beta$-TCVAE on dSprites, Cars3d, Shapes3d, Noisy-dSprites, and Color-dSprites for 1800 settings. These settings include different random seeds and hyperparameters, refer to Appendix G for the details. We only show the first three datasets here. For more results, please refer to Appendix I.}
\label{tab:quant}
\end{table*}

\subsection{Quantitative Evaluations}
This section performs quantitative evaluations on the datasets and models introduced with different random seeds and different hyperparameters. Then, we evaluate the performance of the Original and \emph{Groupified} VAEs in terms of several popular metrics: BetaVAE score~\citep{higgins2016beta}, DCI disentanglement~\cite{eastwood2018framework} (DCI in short), MIG~\citep{chen2018isolating}, and FactorVAE score~\citep {kim2018disentangling}. We assign three or four hyperparameter settings for each model on each dataset. We run it with ten random seeds for each hyperparameter setting to minimize the influence of random seeds. Therefore, we totally run $((3\times10\times3 + 10\times3\times3)\times 2)\times 5 = 1800$
models. We evaluate each metric's mean and variance for each model on each dataset to demonstrate the effectiveness of our method. As shown in Table \ref{tab:quant}, these \emph{Groupified} VAEs have better performance (numbers marked bold in Table \ref{tab:quant}) than the Original VAEs on almost all the cases. 

\begin{figure}[htbp!]
\centering
\begin{minipage}[t]{0.48\textwidth}
\centering
\begin{tabular}{c@{\hspace{0.5em}}c}
{\includegraphics[width=0.47\linewidth]{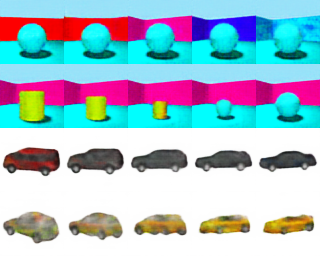}} &
{\includegraphics[width=0.47\linewidth]{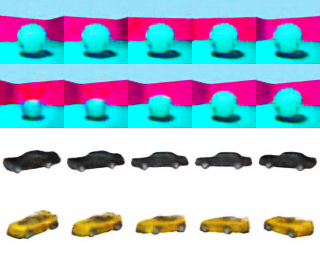}} \\
(a) Original & (b) Groupified \\
\end{tabular}
\caption{Visual traversal comparison between Original and \emph{Groupified} $\beta$-TCVAE. The traversal results of \emph{Groupified} VAEs are less entangled.}
\label{fig:qual_compare}
\end{minipage}
\hspace{0.2em}
\begin{minipage}[t]{0.48\textwidth}
\centering
\begin{tabular}{c@{\hspace{0.2em}}c}
{{\rotatebox{90}{\hspace{2mm}{\scriptsize Original}}}}&{\includegraphics[width=0.93\linewidth]{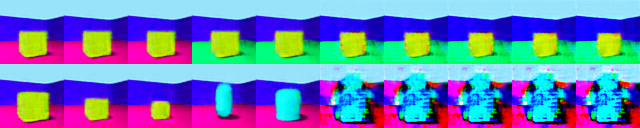}} \\
{{\rotatebox{90}{\hspace{1mm}{\scriptsize Groupified}}}} &{\includegraphics[width=0.93\linewidth]{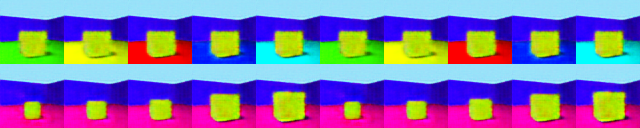}} \\
\end{tabular}
\caption{Traversal results of two factors (floor color, scale) of Original and \emph{Groupified} $\beta$-TCVAE. The traversal results of \emph{Groupified} VAEs are cyclic.}
\label{fig:cyclic}
\end{minipage}
\vspace{-1em}
\end{figure}

On Shapes3d, the \emph{Groupified} VAEs outperform the Original ones on all the metrics except for BetaVAE scores, suggesting some disagreement between BetaVAE scores and other metrics. Similar disagreement is also observed between the variances of MIG and other metrics on Cars3d. Note that the qualitative evaluation in Appendix J shows that the disentanglement ability of \emph{Groupified} VAEs is better on Shapes3d and Cars3d.

\subsection{Qualitative Evaluations}
We qualitatively show the \emph{Groupified} VAEs achieve better disentanglement than the Original ones. As shown in Figure~\ref{fig:qual_compare}, the traversal results of \emph{Groupified} $\beta$-TCVAE on Shape3d and Car3d are less entangled. For more qualitative evaluation, please refer to Appendix J. To verify that the \emph{Groupified} VAEs learn a cyclic representation space (where $n=10$), we provide the traversal results of [0,18] with a step of 2 for both the \emph{Groupified} and Original $\beta$-TCVAE on Shape3d in Figure \ref{fig:cyclic}. We observe that the traversal results of \emph{Groupified} VAEs are of high quality with a period of $10$ (equal to $n$). However, the Original VAEs generate low-quality images without cyclicity. For the comparison of the results on CelebA (real-world datasets), please see appendix J.

\subsection{Visualization of the Learned Representation Space}
\label{Sec:latent_space}
To understand how our theoretical framework helps the existing VAE-based models to improve the disentanglement ability, we take dSprites as an example, visualize the learned representation space, and show the typical score distributions of the metrics. First, we visualize the space spanned by the three most dominant factors (x position, y position, and scale). 

\begin{figure}[htbp!]
\centering
\subfigure[BetaVAE score]{
\vspace{-1.5em}
\begin{minipage}[t]{0.24\linewidth}
\centering
\includegraphics[width=1.3in]{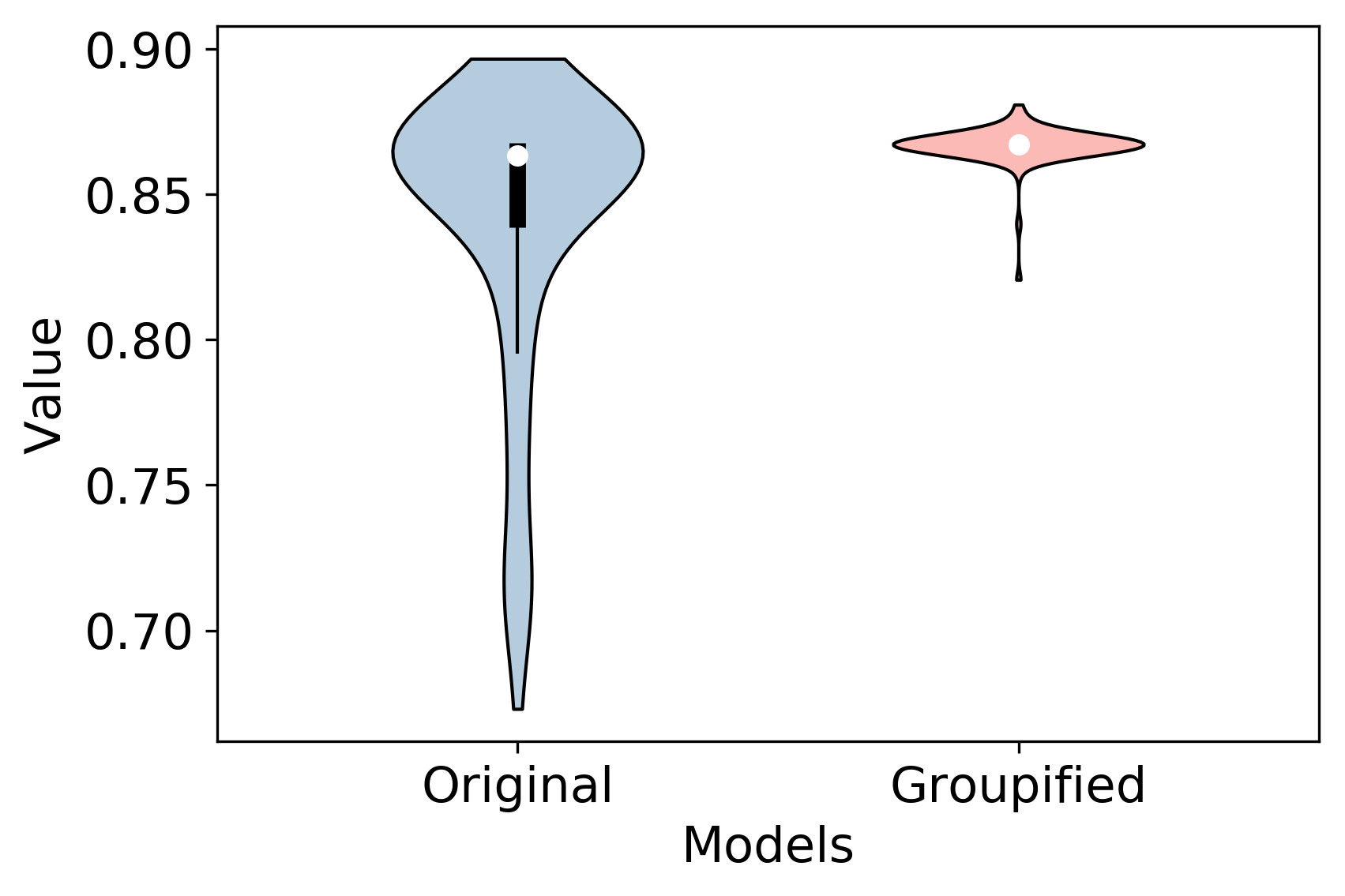}
\end{minipage}%
}%
\subfigure[DCI disentanglement]{
\begin{minipage}[t]{0.24\linewidth}
\centering
\includegraphics[width=1.3in]{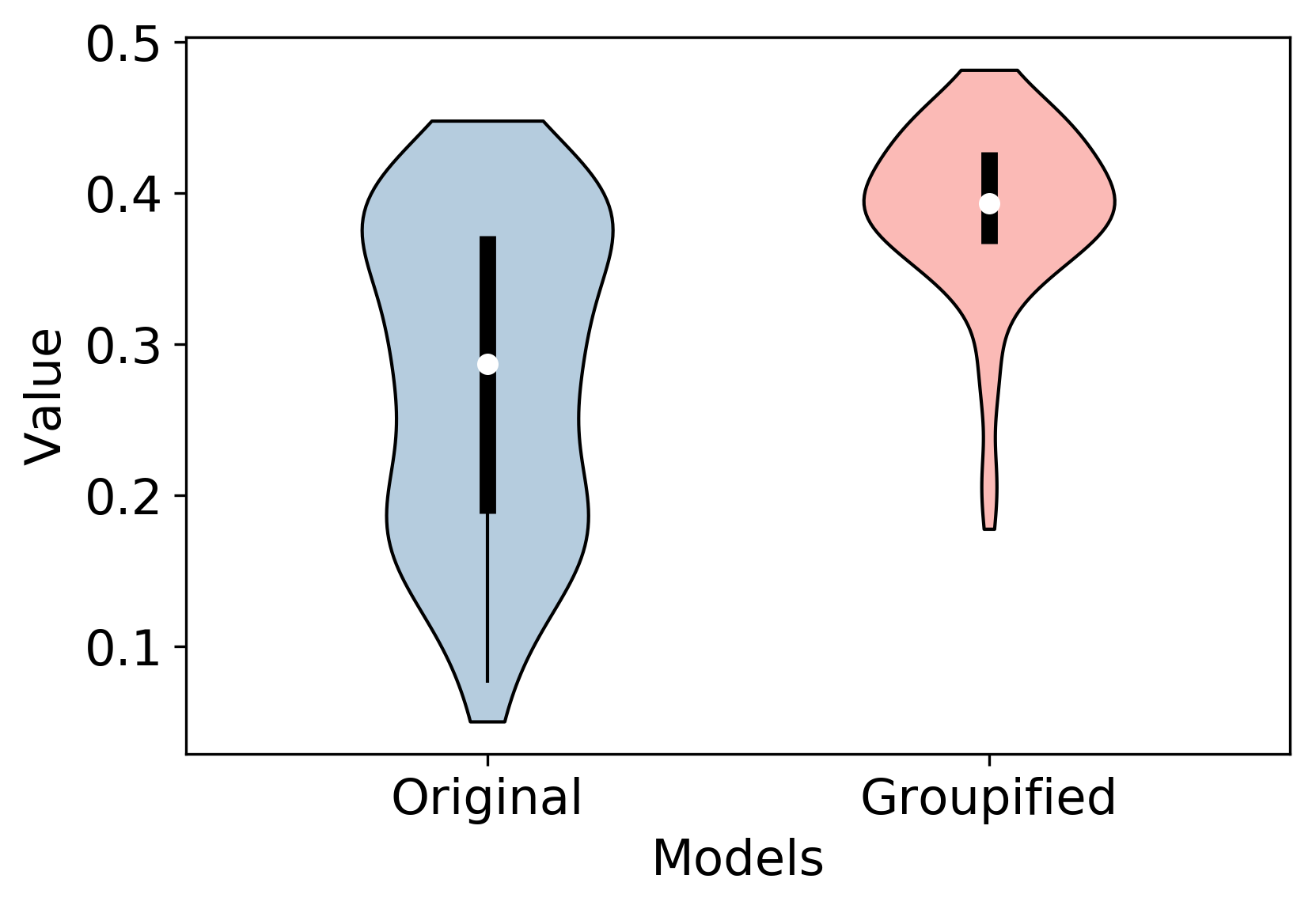}
\end{minipage}%
}%
\centering
\subfigure[MIG]{
\begin{minipage}[t]{0.24\linewidth}
\centering
\includegraphics[width=1.3in]{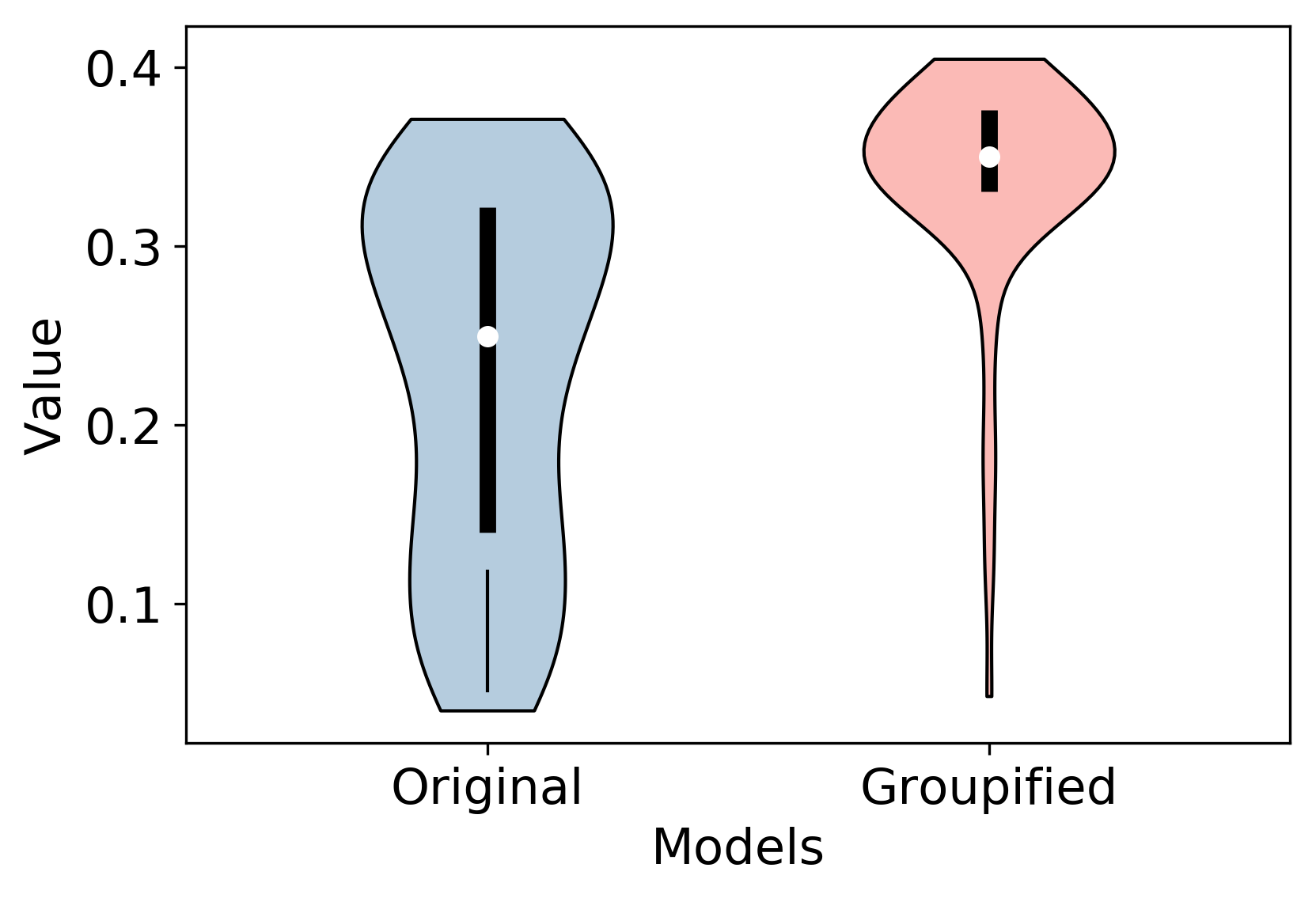}
\end{minipage}%
}%
\subfigure[FactorVAE score]{
\begin{minipage}[t]{0.24\linewidth}
\centering
\includegraphics[width=1.3in]{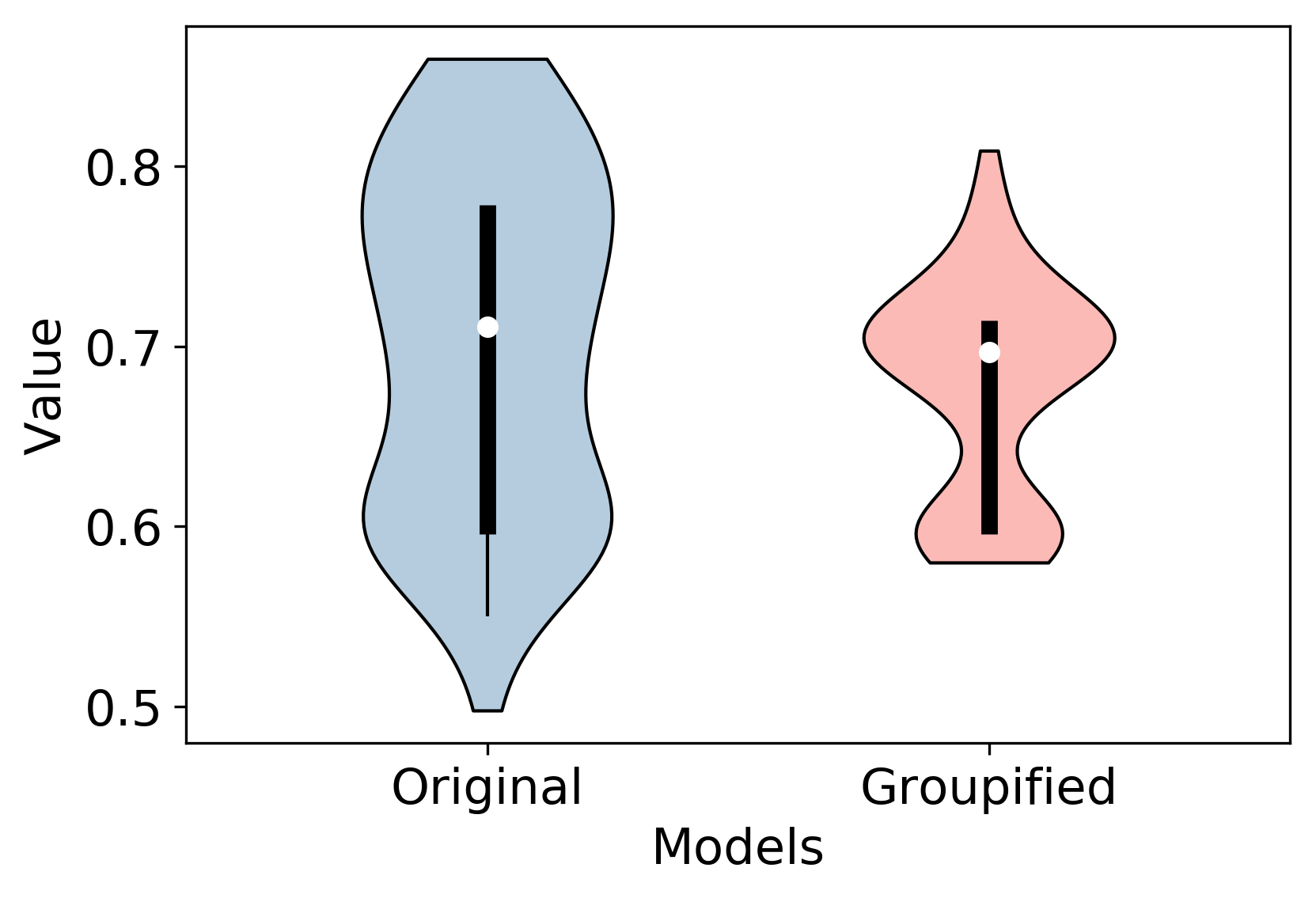}
\end{minipage}%
}%
\centering
\caption{Performance distribution of Original and \emph{Groupified} AnnealVAE on dSprites (demonstrated by the Violin Plot~\citep{hintze1998violin}). Variance is
due to different hyperparameters and random seeds.
We observe that \emph{Groupified} AnnealVAE improves the average performance with smaller variance in terms of BetaVAE score (a), DCI disentanglement (b), and MIG (c), and has a comparable mean performance with smaller variance in terms of FactorVAE score (d).}
\vspace{-1em}
\label{fig:dist}
\end{figure}

\begin{figure*}[htbp!]
\centering
\subfigure[$C=10$, Groupified]{
\begin{minipage}[t]{0.24\linewidth}
\centering
\includegraphics[width=1.3in]{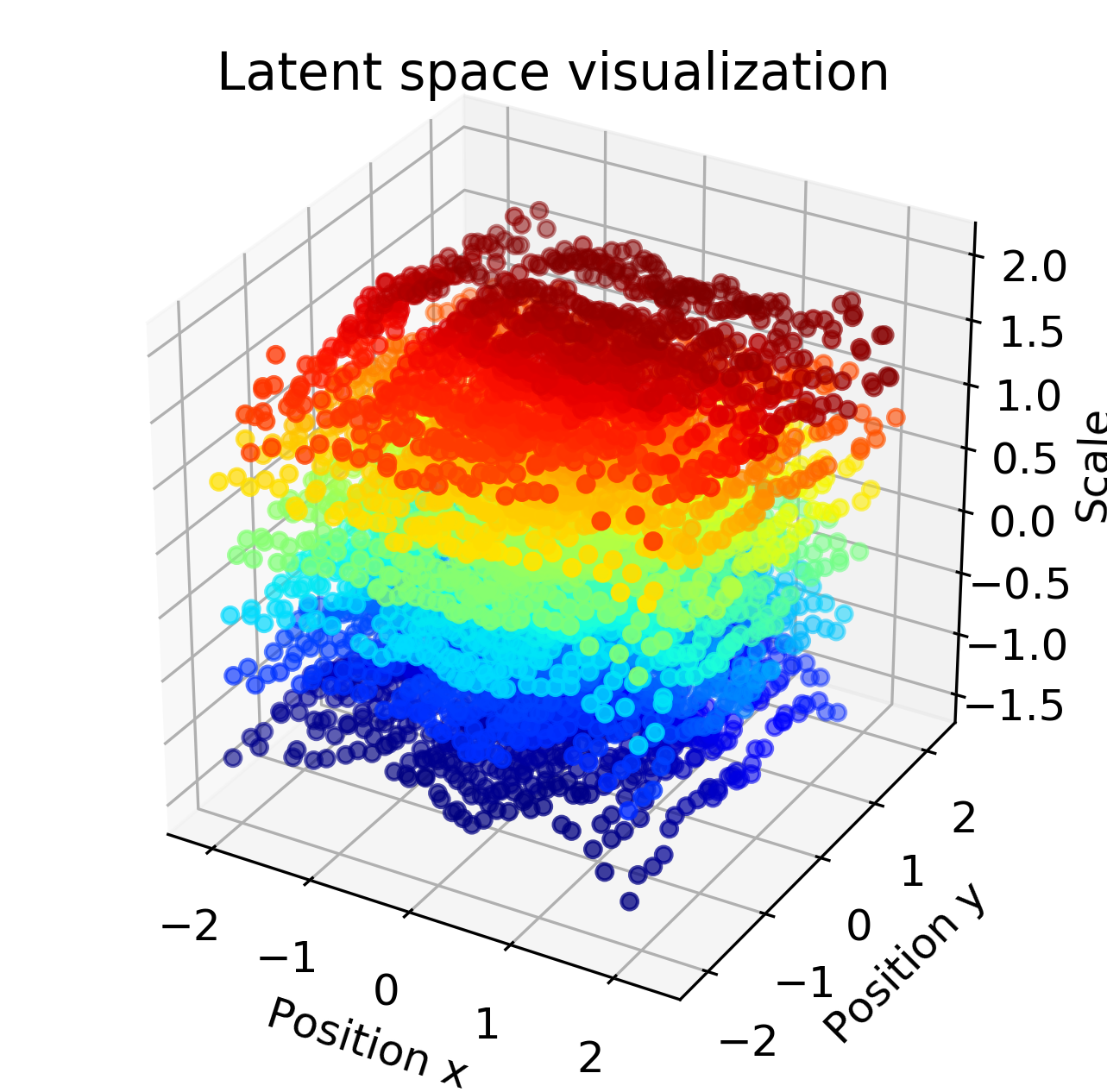}
\end{minipage}%
}%
\subfigure[$C=20$, Groupified]{
\begin{minipage}[t]{0.24\linewidth}
\centering
\includegraphics[width=1.3in]{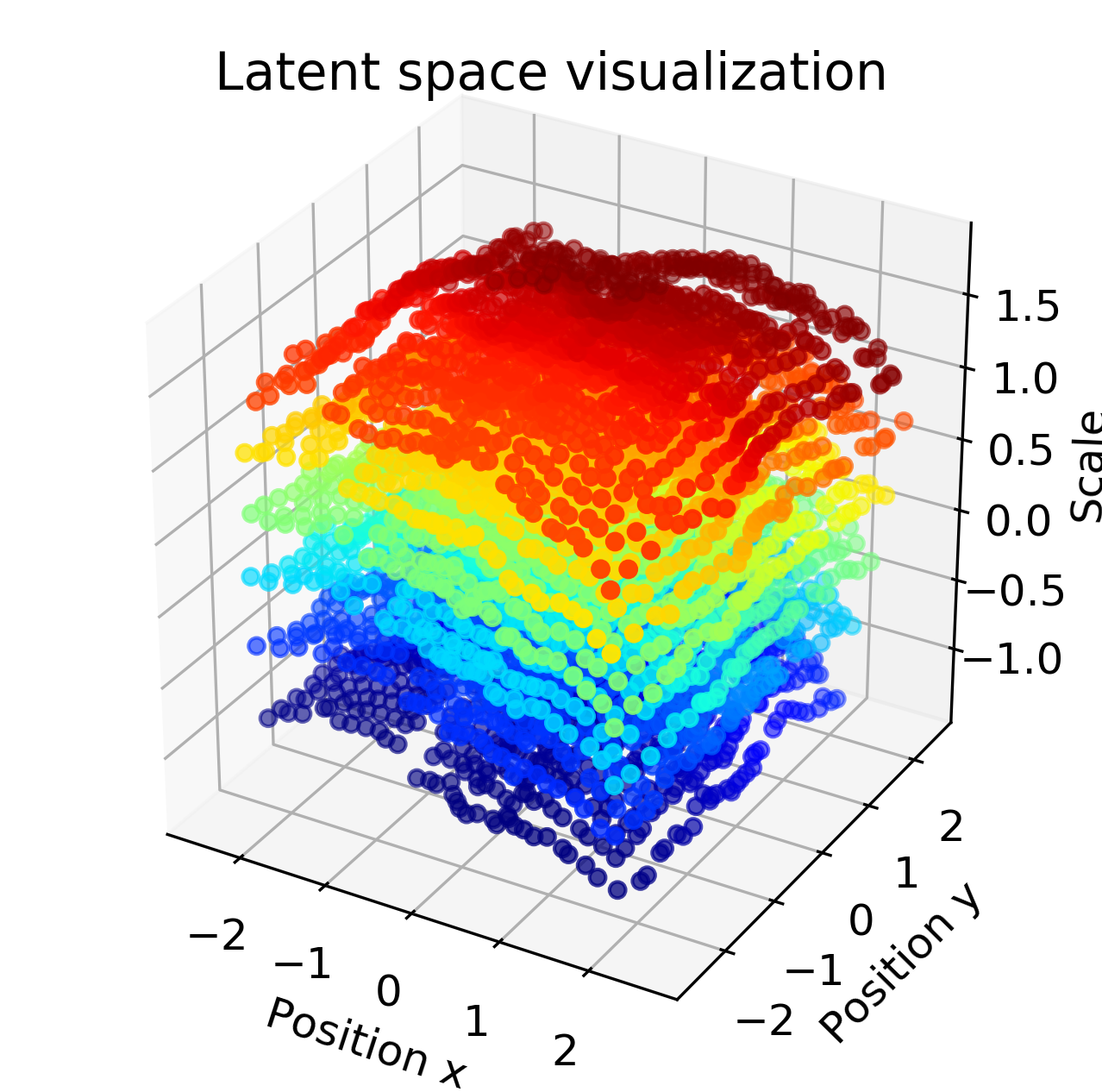}
\end{minipage}%
}%
\subfigure[$C=25$, Groupified]{
\begin{minipage}[t]{0.24\linewidth}
\centering
\includegraphics[width=1.3in]{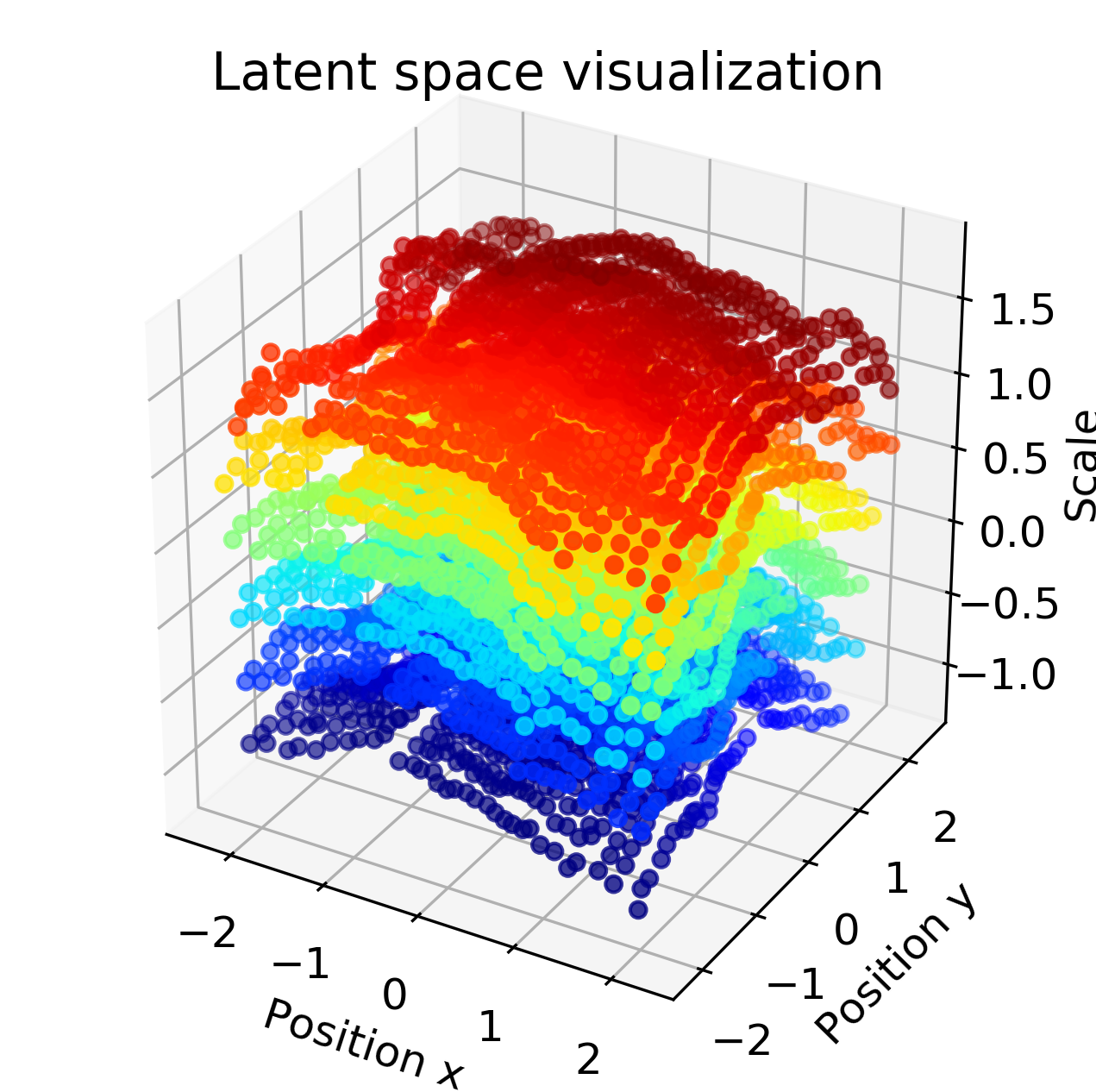}
\end{minipage}%
}%
\subfigure[$C=30$, Groupified]{
\begin{minipage}[t]{0.24\linewidth}
\centering
\includegraphics[width=1.3in]{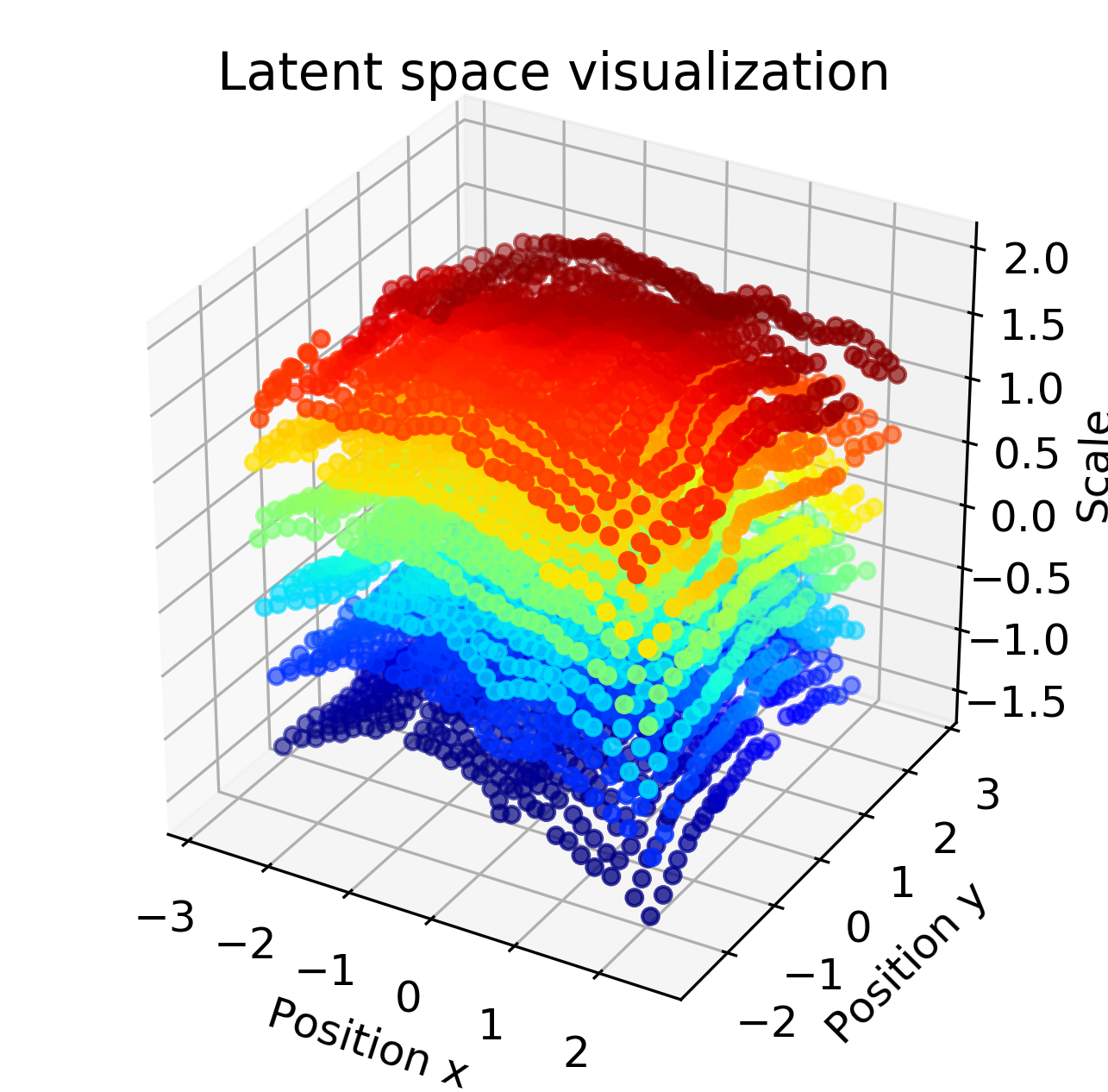}
\end{minipage}%
}%
\centering

\centering
\subfigure[$C=10$, Original]{
\begin{minipage}[t]{0.24\linewidth}
\centering
\includegraphics[width=1.3in]{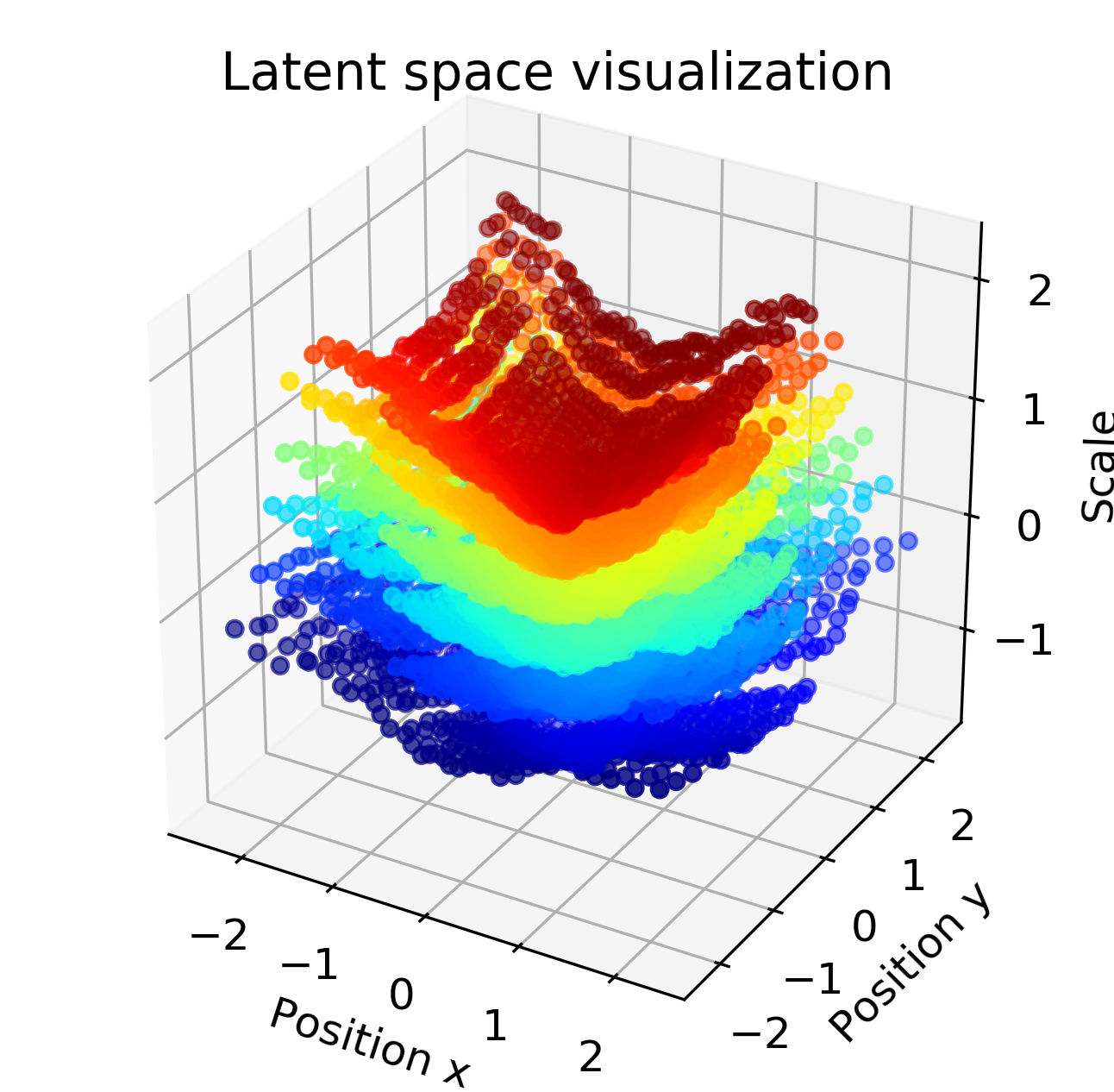}
\end{minipage}%
}%
\subfigure[$C=20$, Original]{
\begin{minipage}[t]{0.24\linewidth}
\centering
\includegraphics[width=1.3in]{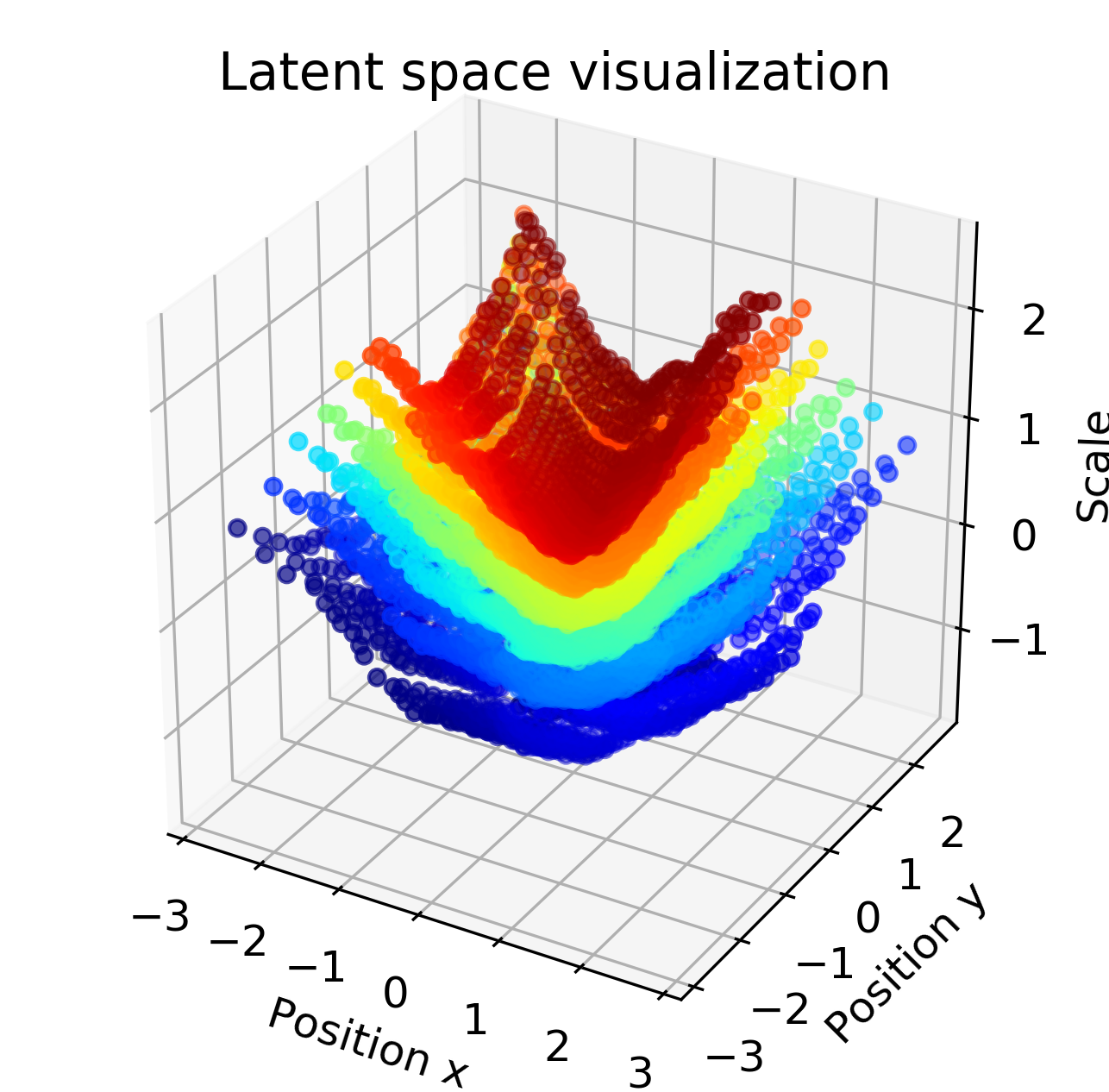}
\end{minipage}%
}%
\centering
\subfigure[$C=25$, Original]{
\begin{minipage}[t]{0.24\linewidth}
\centering
\includegraphics[width=1.3in]{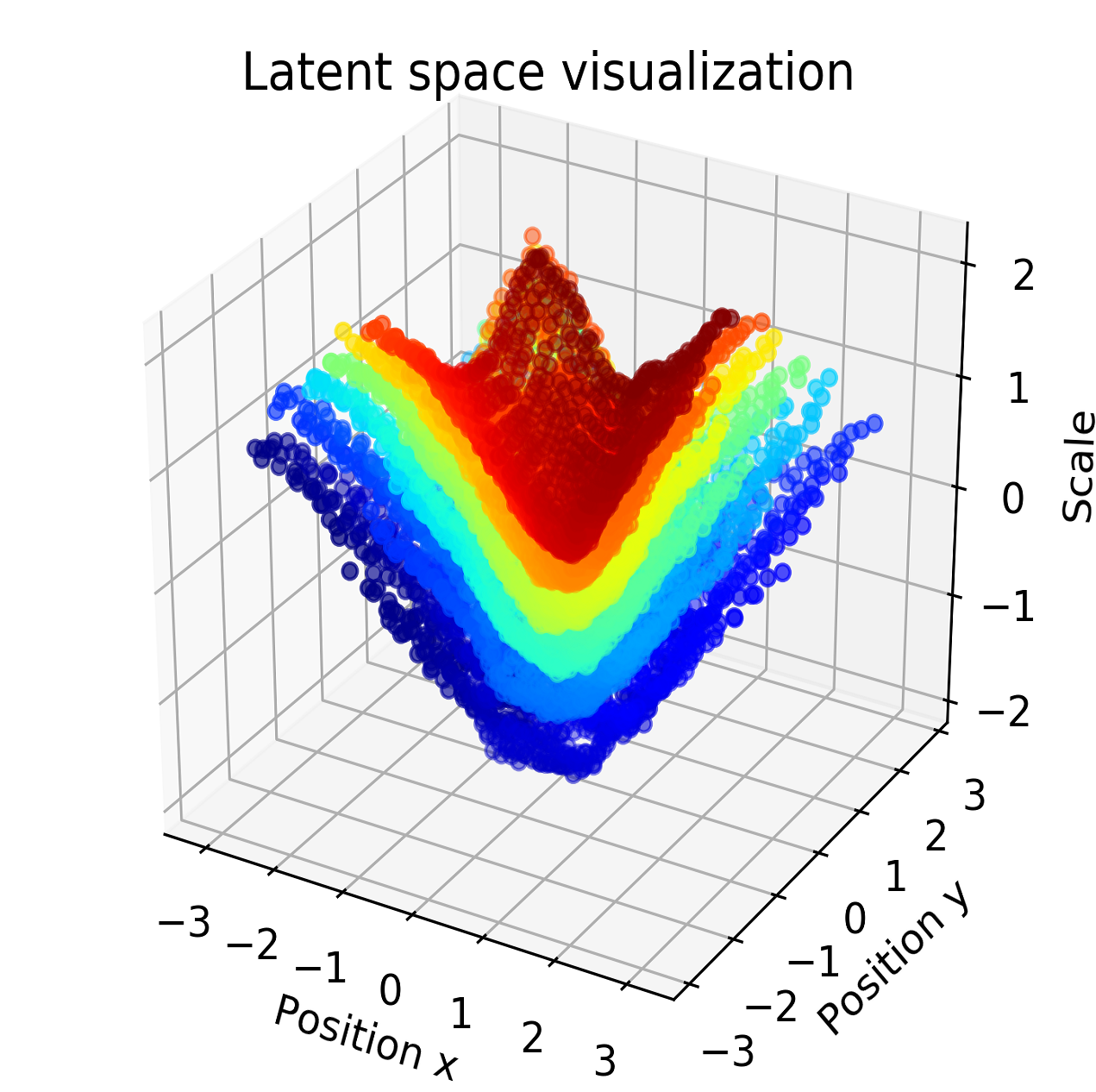}
\end{minipage}%
}%
\subfigure[$C=30$, Original]{
\begin{minipage}[t]{0.24\linewidth}
\centering
\includegraphics[width=1.3in]{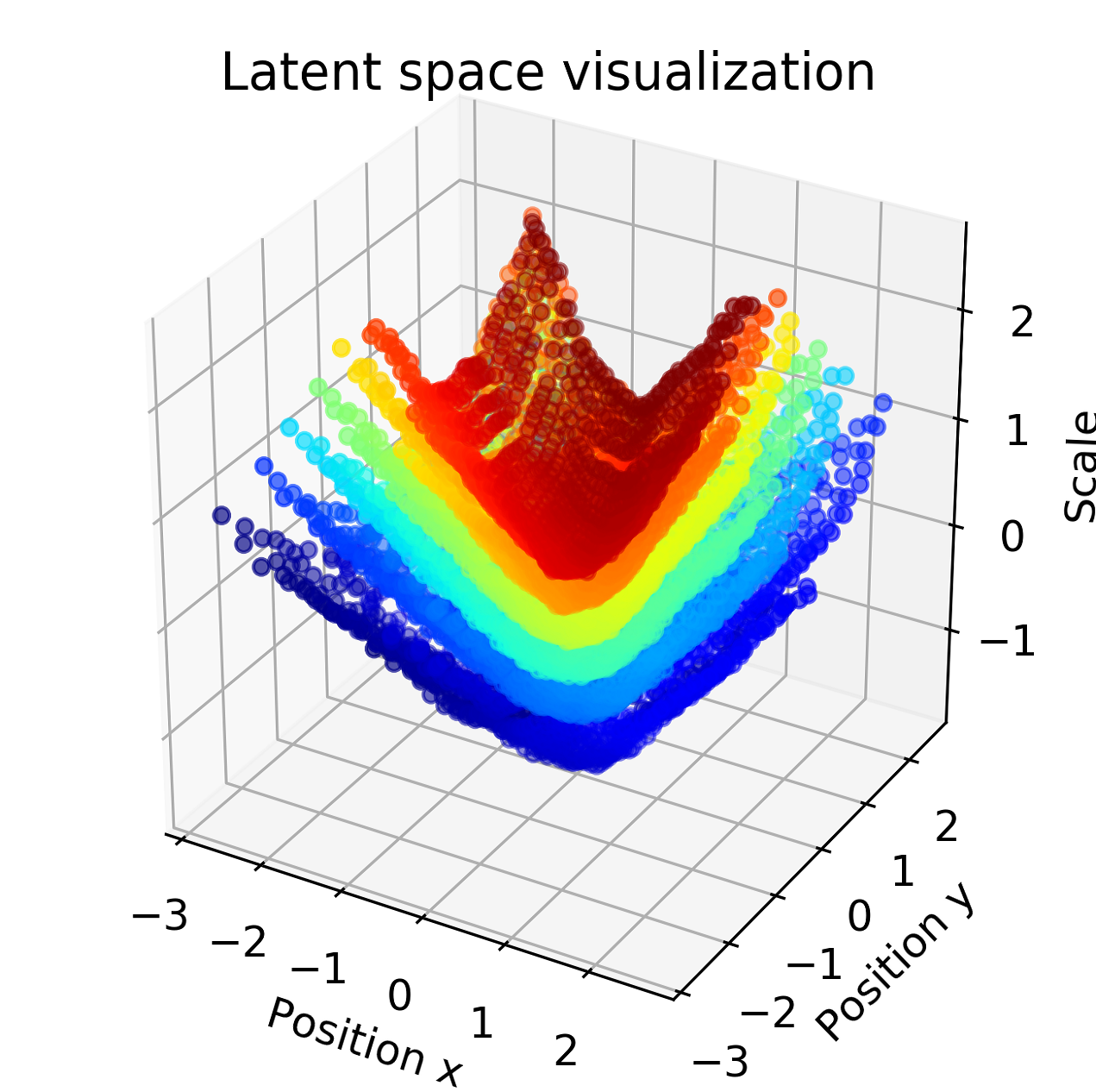}
\end{minipage}%
}%
\centering

\centering
\vspace{-0.5em}
\caption{The representation space spanned by the learned factors by Original (bottom row) and \emph{Groupified} AnnealVAE (top row). The position of each point is the disentangled representation of the corresponding image. An ideal result is all the points form a cube and color variation is continuous. The increase of $C$ (a hyperparameter of AnnealVAE) results in a collapse of representation space of the Original VAE. The collapse is suppressed by the Isomorphism Loss, which leads to better disentanglement.}
\label{fig:space}
\end{figure*}

As shown in Figure \ref{fig:space} (for more results, please refers to Appendix L), the spaces learned by the Original VAEs collapse, while the spaces of the \emph{Groupified} VAEs only bend a little bit. The main reason is that the Isomorphism Loss, serving as a self-supervision signal, suppresses the representation space distortion and encourages the disentanglement of the learned factors. As Figure \ref{fig:dist} shows, the \emph{Groupified} VAEs consistently achieve better mean performance with smaller variances. The isomorphism reduces the search space of the network so that the \emph{Groupified} VAEs converge to the ideal disentanglement solution.

\begin{table*}[htbp!]
\centering

\resizebox{\textwidth}{!}{
\begin{tabular}{c c c>{\columncolor{mygray}}cc cc>{\columncolor{mygray}}c}
\toprule
 & \multirow{2}*{Original} & \multicolumn{3}{c}{Groupified} & \multicolumn{3}{c}{Factor Size $n=10$}\\ 
\cmidrule(lr){3-8}
 &  & $n=5$ & $n=10$ & $n=15$ & w/o Abel & w/o Order & Groupified \\
 \midrule
 DCI & $ 0.27 \pm 0.10 $ & $ 0.34 \pm 0.062 $  & $ \bm{0.38} \pm \bm{0.055} $  & $ \bm{0.38} \pm 0.064 $  & $ 0.28 \pm 0.11 $ & $ 0.34 \pm 0.056 $ & $\bm{0.38} \pm \bm{0.055}$\\
 \bottomrule
\end{tabular}}
\caption{Ablation study on the factor size $n$ and Isomorphism Loss. DCI disentanglement is listed (mean $\pm$ std).}
\label{tab:abla}
\vspace{-1 em}
\end{table*}

\subsection{Ablation Study}
We perform an ablation study on the assumed total number of possible values for a factor (factor size) $n$, Abel Loss $\mathcal{L}_{a}$, and Order Loss $\mathcal{L}_{o}$. We take the AnnealVAE trained on dSprites as an example. We only consider the DCI disentanglement metric here. We investigate the influence of factor size $n$. Besides, to evaluate the effectiveness of the two constraints, the models with the Abel Loss alone or Order Loss alone added are also evaluated. In this setting, we fix $n$ to 10. We compute the mean and variance of the performance for 30 settings of hyperparameters and random seeds. Table \ref{tab:abla} shows that the isomorphism plays a role of cycle consistency in the representation space, leading to better disentanglement. The performance is robust to the factor size $n$, as the models learn to adapt to different $n$ in the training process.  The models with only the Abel Loss or Order Loss applied have improved performance compared to the originals. The former (Abel Loss) performs better than the latter, suggesting that commutativity plays a more important role. Note that the number of factors $m$ can be learned and is not a hyperparameter. See Appendix F for details. $\gamma_I$ is empirically set to $1$.

\section{Conclusion}
In this paper, we have opened the possibility of applying group-based definition to \emph{unsupervised} disentanglement by proposing a theoretical framework. The group structure and model constraint in the framework are effective for existing VAE-based \emph{unsupervised} disentanglement methods. In addition, by establishing the feasibility of learning the representation conforming to the definition in \emph{unsupervised} settings, we have exhibited the consistently better mean performance with lower variance attributed to the definition. We believe our work constitutes a promising step towards \emph{unsupervised} disentanglement with theoretical guarantee. As to the limitation, we only provide a necessary condition for the \emph{data constraint}, as a result, we can not address the unidentifiability problem. Tackling the unidentifiability problem with the group-based definition is beyond the scope of this work, we will leave it as future work. In addition, a natural extension of our framework is to use \emph{lie group}~\cite{hall2015lie} (which is also a manifold) to extend our framework.

\bibliographystyle{iclr2022_conference}
\bibliography{merged_version}
\end{document}